\documentclass[twoside,leqno,twocolumn]{article}

\usepackage{siamproceedings}

\usepackage[T1]{fontenc}
\usepackage{amsfonts}
 
\usepackage{amsmath}
\usepackage{mathrsfs}
\usepackage{bm}
\usepackage{xcolor}
\usepackage{colortbl}
\definecolor{secondbest}{gray}{0.9}
\usepackage{graphicx}
\usepackage{subcaption}
\usepackage{epstopdf}
\ifpdf
  \DeclareGraphicsExtensions{.pdf,.png,.jpg,.eps}
\else
  \DeclareGraphicsExtensions{.eps}
\fi
 
\usepackage{booktabs}
\usepackage{fancyhdr}
\usepackage{multirow}
\usepackage{makecell}
\usepackage{tabularx}
\usepackage{xcolor}
 
\usepackage{enumitem}
\usepackage{xspace}
\usepackage{multicol}
\usepackage{algorithmic}
\usepackage{algorithm}
 
\usepackage{threeparttable}
\usepackage{wrapfig}
\usepackage{tcolorbox}
\tcbuselibrary{breakable}
\usepackage{float}
\usepackage{listings}

\usepackage{amsopn}

\begin{document}

\newcommand\relatedversion{}


\title{\Large CRAFT: LLM-Based Iterative Refinement for Temporal Reasoning over Clinical Narratives}
 
\author{Chengyang He\thanks{Stevens Institute of Technology, Hoboken, NJ, 07030, USA (\email{che14@stevens.edu},
\email{tarif1@stevens.edu},
\email{yue.ning@stevens.edu}, 
\email{pwang44@stevens.edu})}
    \and Tahreem Arif\footnotemark[1]
    \and Marko Zivkovic\thanks{Genesis Research Group, Hoboken, NJ, 07030,USA (\email{marko.zivkovic@genesisrg.com})}  
    \and Lijing Wang\thanks{New Jersey Institute of Technology, Newark, NJ, 07102, USA 
  (\email{lijing.wang@njit.edu}}
  \and Yue Ning\footnotemark[1]
  \and Ping Wang\footnotemark[1]
  }

\date{}

\maketitle


\fancyfoot[R]{\scriptsize{Copyright \textcopyright\ 2026 by SIAM\\
Unauthorized reproduction of this article is prohibited}}





\begin{abstract}

Understanding the temporal progression of symptoms in clinical narratives
is critical for disease monitoring, safety surveillance, and causality
assessment. Clinical narratives, however, rarely provide explicit temporal
anchors. Current approaches to temporal information reasoning focus
predominantly on pairwise relation classification across multi-visit and 
timestamp-rich records, leaving the reconstruction of structured symptom
trajectories from individual anchor-sparse reports largely unaddressed.
We propose CRAFT, an LLM framework that pairs a generator with a
constraint-based verifier to iteratively produce and refine stage-wise
symptom timelines through targeted feedback. We conduct evaluation on MedTempo, a new benchmark
of 5,347 vaccine adverse-event narratives spanning three COVID-19 vaccine
types, with expert-validated temporal stage annotations for 3,166 reports.
Experiments across four LLM backbones demonstrate that CRAFT consistently
improves temporal ordering accuracy, with ablation analysis isolating the
contribution of generator and verifier components across model capability
levels.

\end{abstract}

\section{Introduction.}

Clinical temporal reasoning, recovering a stage-wise trajectory of clinical events from narrative text, is central to disease progression modeling, treatment outcome monitoring, and safety signal detection~\cite{cui-etal-2025-timer,olex-mcinnes-2021-review}. However, constructing such orderings from free text remains labor-intensive. Temporal cues in these narratives are frequently implicit or expressed relative to other events rather than anchored to fixed dates, and
co-mentions, restatements, and status updates further obscure the true chronological order of symptoms~\cite{zhou-parsons-hripcsak-2008-temporal,alfattni-peek-nenadic-2020-review}.
 
A large body of work on temporal information reasoning has focused on identifying events and temporal expressions and predicting pairwise relations to derive a global ordering~\cite{leeuwenberg-moens-2018-temporal,savova2009towards}, with recent efforts expanding to exhaustive relation coverage and complex temporal fact extraction~\cite{alsayyahi-batista-navarro-2023-timeline,chen-etal-2024-tsdre}. In the clinical domain, however, progress has been bottlenecked by limited benchmark diversity, with most work concentrated on a small number of corpora and relation inventories~\cite{alfattni-peek-nenadic-2020-review,su-howard-bethard-2025-review}. Recent LLM-based approaches to clinical temporal reasoning further assume multi-visit timelines, timestamp-linked supervision, or constrained data settings~\cite{andrew-etal-2025-llm-trel,cui-etal-2025-timer,yu-stidham-vydiswaran-2023-pipeline}. As a result, standardized methods and benchmarks for ordering symptom progressions within single-report, anchor-sparse clinical narratives remain underexplored.
 
To address this gap, we propose \textbf{CRAFT} (\textbf{C}linical \textbf{R}efinement with \textbf{A}daptive \textbf{F}eedback for \textbf{T}emporal ordering), a generator--verifier framework that models temporal trajectory reconstruction as an iterative structured prediction task under weak temporal anchoring, where candidate trajectories are refined via constraint-based feedback. Inspired by iterative refinement paradigms such as Self-Refine~\cite{selfrefine}, CRAFT introduces a structured trajectory representation and a task-specific verification mechanism tailored to temporal ordering, and can be instantiated with different generator and verifier configurations. In this work, we instantiate CRAFT as \textbf{CRAFT-Full}, which pairs a full-regeneration generator with a multi-criterion additive verifier; we additionally define two baselines (\textbf{PIVOT}, \textbf{GUIDE}) and two ablations (\textbf{CRAFT-G}, \textbf{CRAFT~w/o~V}) to isolate the contribution of the generator and verifier components respectively.
 
To enable rigorous evaluation, we introduce \textbf{MedTempo} (\textbf{Med}ical \textbf{Temp}oral \textbf{O}rdering Benchmark), a benchmark for temporal progression reconstruction from medical free text. MedTempo contains \textbf{5,347} narrative reports spanning three vaccine types, each consisting of a single report per patient with no explicit absolute time anchor and paired with a provided symptom list. Our benchmark task focuses on the \textbf{3,166} reports that exhibit temporal evidence of distinct symptom progression, for which we provide expert-validated stage-wise ordering annotations. The remaining reports contain no temporal progression and are retained in the dataset release to support future work on temporal-evidence identification, but fall outside the scope of the current benchmark evaluation. Our primary contributions are as follows:

\begin{itemize}
    \vspace{-1mm}
    \item We propose CRAFT, a generator--verifier framework for iterative temporal reasoning refinement under weak temporal anchoring, with controlled baselines and ablations that isolate generator and verifier contributions.
    \item We introduce MedTempo, an expert-annotated benchmark for evaluating structured temporal trajectories over anchor-sparse clinical narratives.
    \item We conduct extensive experiments across four LLMs, revealing distinct refinement behaviors tied to model capability and verifier calibration.
    \vspace{-1mm}
\end{itemize}
Figure~\ref{fig:overview} provides a schematic overview of the full pipeline, from dataset construction through iterative refinement and post-hoc evaluation.

\section{Related Work.}

Temporal information extraction has been shaped by the TimeML annotation framework~\cite{pustejovsky-etal-2003-timeml}, which introduced a markup language for events, temporal expressions, and their relations. The TempEval shared tasks~\cite{verhagen-etal-2007-tempeval} established standardized evaluation for pairwise temporal relation classification, with systems progressing from rule-based and CRF/SVM approaches to neural methods. A common paradigm is to predict pairwise relations and then induce a globally consistent ordering~\cite{leeuwenberg-moens-2018-temporal,cheng-miyao-2018-inducing,dct}, while more recent benchmarks emphasize richer annotation for event ordering~\cite{alsayyahi-batista-navarro-2023-timeline} and LLM-based strategies for temporally grounded fact extraction~\cite{chen-etal-2024-tsdre}. Transformer-based methods have become the dominant paradigm, as surveyed in~\cite{su-howard-bethard-2025-review}. However, these efforts typically center on local relation correctness rather than the end-to-end reconstruction of an ordered, grouped trajectory under weak anchoring.
 
In the clinical domain, the i2b2 2012 challenge~\cite{sun-etal-2013-i2b2} brought temporal relation extraction to clinical discharge summaries, followed by the Clinical TempEval shared tasks~\cite{bethard-etal-2017-clinical-tempeval} on the THYME corpus~\cite{styler-etal-2014-thyme}, where best-performing systems evolved from CRF/SVM classifiers to LSTMs~\cite{tourille-2018-thesis}. Surveys document persistent difficulties including implicit time anchors, inter-sentence relations, and the gap between relation-level extraction and usable patient timelines~\cite{olex-mcinnes-2021-review,alfattni-peek-nenadic-2020-review}. More recent work applies neural end-to-end methods to established clinical corpora~\cite{miller-etal-2023-end-to-end}, and LLM-based approaches have begun to examine prompting and fine-tuning for clinical temporal relation extraction~\cite{he-etal-2024-prompting-ctre,andrew-etal-2024-evaluating,andrew-etal-2025-llm-trel,yuan-etal-2023-zeroshot-tre} as well as timeline extraction from medical case reports~\cite{wang-weiss-2025-relative-timeline}. However, these approaches typically require longitudinal records spanning multiple visits or rely on structured temporal metadata~\cite{cui-etal-2025-timer,yu-stidham-vydiswaran-2023-pipeline}. In contrast, CRAFT operates on single-report clinical narratives where temporal cues are sparse or implicit. To support evaluation in this underexplored setting, MedTempo provides expert-annotated temporal trajectory benchmarks derived from real-world adverse event narratives.

\begin{table}[!t]
\centering
\small
\setlength{\tabcolsep}{2.8pt}
\renewcommand{\arraystretch}{1.05}
\caption{Descriptive statistics for MedTempo by vaccine type. Text Len.\ denotes clinical narrative length in words; \#Stages denotes
number of temporal stages.}
\label{tab:vaers_statistics_detailed}
\resizebox{0.5\textwidth}{!}{
\begin{tabular}{ll rrr rrr rrr}
\toprule
& & \multicolumn{3}{c}{\textbf{MedTempo}} & \multicolumn{3}{c}{\textbf{MedTempo-T}} & \multicolumn{3}{c}{\textbf{MedTempo-NT}} \\
\cmidrule(lr){3-5}\cmidrule(lr){6-8}\cmidrule(lr){9-11}
\textbf{Vaccine} & \textbf{Metric} & \textbf{Med} & \textbf{Min} & \textbf{Max}
                                   & \textbf{Med} & \textbf{Min} & \textbf{Max}
                                   & \textbf{Med} & \textbf{Min} & \textbf{Max} \\
\midrule
\multirow{3}{*}{\shortstack[l]{\textbf{Pfizer}\\(1789/1019/770)}}
& Text Len.   & 102 & 11 & 2319 & 109 & 12 & 1638 & 81 & 11 & 2319 \\
& \# Symp. &   6 &  4 &   64 &   6 &  4 &   29 &  5 &  4 &   64 \\
& \# Stages    &   2 &  0 &   10 &   3 &  1 &   10 &  0 &  0 &   0 \\
\midrule
\multirow{3}{*}{\shortstack[l]{\textbf{Moderna}\\(1769/983/786)}}
& Text Len.   &  83 & 11 & 1056 &  99 & 13 &  905 & 57 & 11 & 1056 \\
& \# Symp. &   6 &  4 &   30 &   6 &  4 &   30 &  5 &  4 &   29 \\
& \# Stages    &   2 &  0 &    9 &   3 &  2 &    9 &  0 &  0 &    0 \\
\midrule
\multirow{3}{*}{\shortstack[l]{\textbf{Janssen}\\(1789/1164/625)}}
& Text Len.   &  87 & 11 & 1317 & 100 & 12 &  826 & 56 & 11 & 1317 \\
& \# Symp. &   6 &  4 &   43 &   7 &  4 &   43 &  5 &  4 &   33 \\
& \# Stages    &   2 &  0 &   13 &   3 &  2 &   13 &  0 &  0 & 0 \\
\bottomrule
\end{tabular}
}
\end{table}


A parallel line of work applies iterative refinement to structured prediction, most notably the Self-Refine paradigm~\cite{selfrefine}, which iterates over model outputs using self-generated feedback. In the clinical domain, Hein et al.~\cite{hein2025iterative} apply iterative refinement with human-in-the-loop review cycles to improve extraction precision. However, such approaches are not suitable for systematic benchmark evaluation across model tiers, as they conflate model capability with human reviewer effort. CRAFT leverages iterative refinement for structured clinical temporal extraction in a fully automated setting, pairing a generator with a multi-criterion verifier and enabling principled comparison across frontier and open-weight models.
 
\section{Dataset Creation.}
This section describes how MedTempo is constructed from VAERS. We first
introduce the source corpus and the information each report provides. We
then describe the filtering and stratified sampling that reduce the corpus
to reports carrying temporal signal, followed by the annotation
protocol that produces the gold-standard timelines. We close with summary
statistics of the resulting benchmark.

\subsection{VAERS Dataset.}

The Vaccine Adverse Event Reporting System (VAERS), managed jointly by the CDC and FDA, is a passive surveillance database in which healthcare professionals, patients, and manufacturers submit reports of adverse events following immunization~\cite{VAERS}. Each report includes demographics, vaccination details, free-text clinical narratives, and MedDRA-coded symptom lists. We focus on three widely administered COVID-19 vaccines: Pfizer-BioNTech, Moderna, and Janssen, covering reports from 2021 to 2024.

\subsection{Data Sampling.}



\begin{table}[!t]
\centering
\small
\setlength{\tabcolsep}{4pt}
\caption{Annotation agreement and adjudication rates (\%) overall and
by subset. Ann.\ = Annotator; T = MedTempo-T with temporal report only;
NT = MedTempo-NT with no temporal report. Ambiguity exclusions are not
applicable (--) in the T subset by definition.}
\label{tab:agreement_adjudication}
\begin{tabular}{l l c c c}
\hline
\textbf{Category} & \textbf{Metric} & \textbf{MedTempo} & \textbf{T} & \textbf{NT}\\
\hline
\multirow{3}{*}{Agreement}
  & Model--Ann.\,1  & 78 & 82 & 91 \\
  & Model--Ann.\,2  & 73 & 78 & 86 \\
  & Inter-Annotator & 93 & 94 & 94 \\
\hline
\multirow{3}{*}{Adjudication}
  & Accepted        & 80 & 90 & 90 \\
  & Corrected       &  9 & 10 & 10 \\
  & Excluded        & 11 & -- & -- \\
\hline
\end{tabular}
\vspace{+3mm}
\end{table}

Starting from the full VAERS corpus, we applied a multi-stage filtering 
and stratified sampling pipeline. We first removed non-symptom MedDRA
terms by building an exclusion list from three non-clinical System Organ
Classes (\textit{Surgical and medical procedures}, \textit{Social
circumstances}, \textit{Product issues}) combined with human-annotated
non-symptom labels, yielding \textbf{5,601} excluded terms. Reports with
three or fewer distinct symptoms were discarded as they lack sufficient
temporal variation. Reports of ten or fewer words (typically bare symptom
lists without narrative context) were also removed. For reports between
11 and 30 words, we applied rule-based temporal keyword filtering
(relative markers such as \textit{before/after}, duration terms, date
patterns) to retain only those with explicit temporal cues; reports of
30 or more words were retained unconditionally. Finally, stratified
sampling balanced by year and report length was applied per vaccine to
draw \textbf{2,000 records each}, preserving the distribution of the
underlying VAERS corpus. Appendix \ref{appendix:data_sampling} presents
representative examples of reports removed under each criterion.

\subsection{Annotation of Temporal Sequence.}



Temporal timelines were produced via a \textit{human-in-the-loop} three-phase protocol: \textit{GPT-4o mini} \cite{openai2024gpt4o} generated initial stage-ordered timelines; two annotators with a medical NLP background independently reviewed and labeled each sequence; and all disagreements and uncertain cases were resolved through collaborative adjudication. As shown in Table \ref{tab:agreement_adjudication}, human inter-annotator agreement (IAA) was 93\%. After adjudication, 80\% of LLM annotations were accepted without change, 9\% were corrected, and 11\% were excluded for temporal ambiguity, yielding a final corpus of 5,347 records.

\subsection{Dataset Statistics and Analysis.}



\begin{figure}[!tp]
    \centering
    \includegraphics[width=0.5\textwidth]{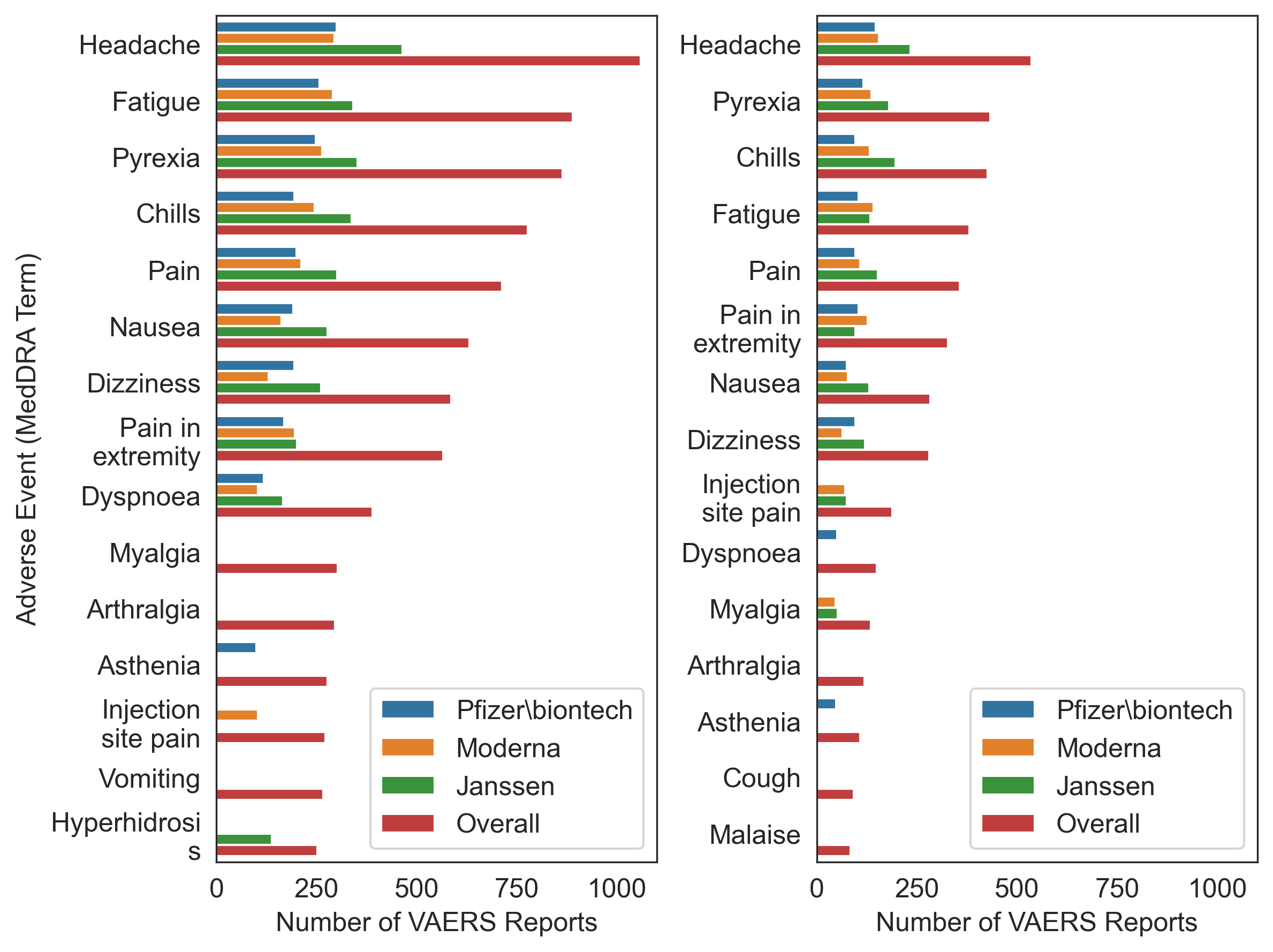}

    \caption{Distribution of the most frequent symptoms (left) and the symptoms that occurred after (right). Each chart displays a total of 15 unique symptoms, representing the top 10 symptoms from each vaccine type and the overall dataset combined.}
    \label{fig:top_15_symptoms}
    \vspace{-5mm}
\end{figure}

Table~\ref{tab:vaers_statistics_detailed} summarizes the 3,166 temporally-evident reports that form the primary benchmark subset \textbf{MedTempo-T}; the remaining 2,181 reports contain no temporal progression and are released separately as \textbf{MedTempo-NT}. Symptom count is consistent across vaccines (median 6), while narrative length varies: Pfizer-BioNTech reports are longest (median 102 words) versus Moderna (83) and Janssen (87). Because symptom count is stable regardless of length, longer reports likely contribute richer contextual cues rather than additional adverse events, providing stronger signals for temporal extraction. Figure~\ref{fig:top_15_symptoms} shows that thermoregulatory symptoms (pyrexia, chills) dominate early stages, while headache and fatigue are the most frequent overall, illustrating the diversity of symptom trajectories in the benchmark.

\section{Method.}
\label{sec:method}
\begin{figure*}[!h]
    \centering
    \includegraphics[trim = 0 0 0 0,clip, width=1\textwidth]{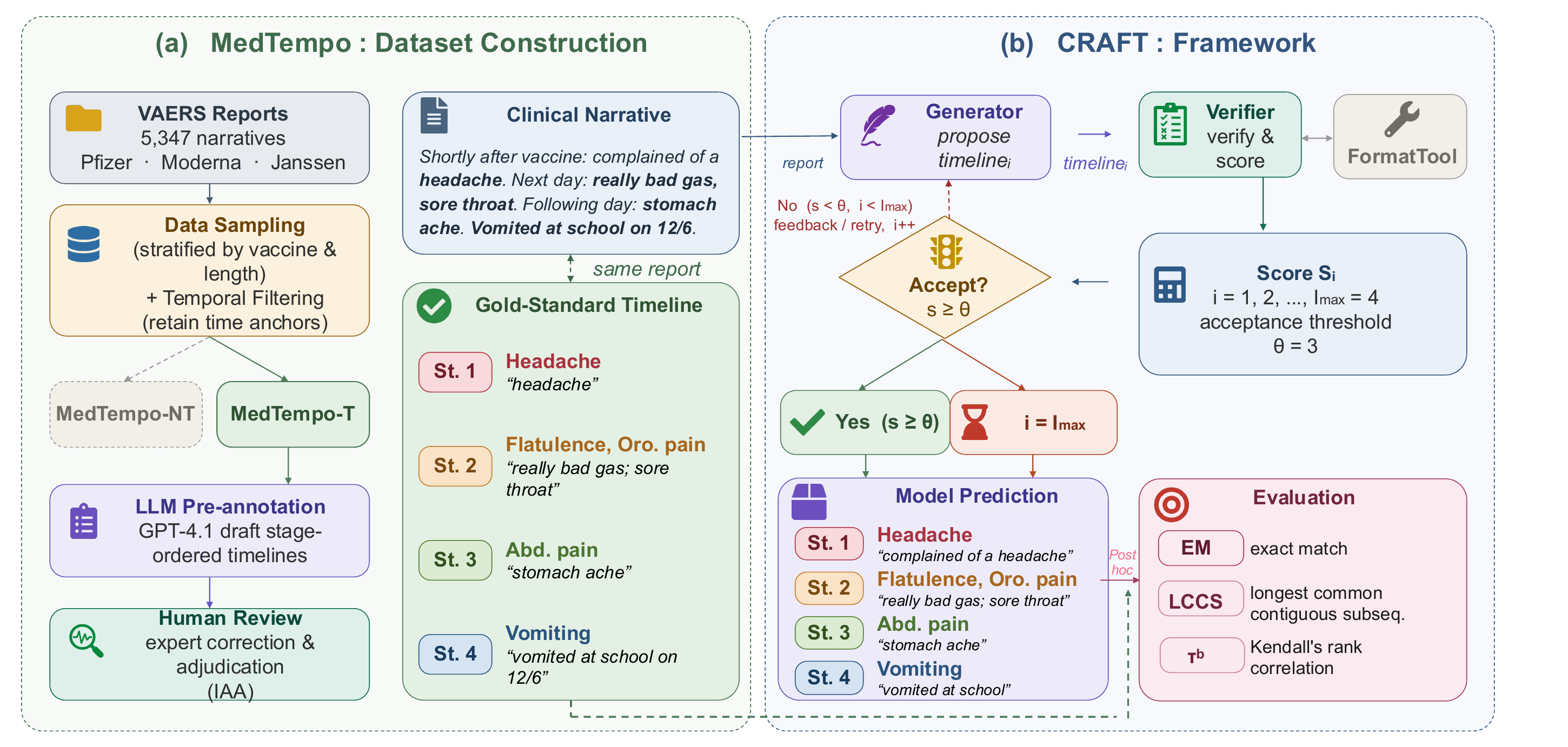}
    \caption{Overview of MedTempo and CRAFT. \textbf{(a)}~Dataset construction pipeline from VAERS narratives to gold-standard stage-ordered timelines. \textbf{(b)}~CRAFT iterative generator--verifier loop, instantiated across four configurations (CRAFT-Full, PIVOT, GUIDE, CRAFT-G) and evaluated on four LLM backbones.
    }
    \label{fig:overview}
    \vspace{-3mm}

\end{figure*}

This section presents CRAFT. We first formalize the ordering task and define
the stage-based representation used throughout the paper, fixing the notation
for reports, findings, and predicted timelines. Then, we describe the framework
itself: the generator that proposes a timeline, the verifier that scores it and
returns feedback, and the loop that connects them. Figure~\ref{fig:overview}
gives an overview of both the dataset pipeline and the framework.

\subsection{Problem Formulation and Temporal Representation.}

For each post-vaccination report \(r\), let \(x_r\) denote its free-text
clinical narrative and \(\mathcal{F}(r) = \{f_1,\dots,f_n\}\) the provided
list of adverse clinical findings (MedDRA Preferred Terms from the VAERS
\texttt{SYMPTOM} fields). We treat \(\mathcal{F}(r)\) as given and do not
perform entity extraction or coding.

Our goal is to predict an explicit temporal ordering over \(\mathcal{F}(r)\)
when the narrative expresses a \emph{temporal progression} (i.e., at least
one new finding appears \emph{after} previously mentioned findings). We exclude
concurrent onset, changes in severity or resolution without new onsets, and
timelines inferable only from durations (e.g., ``finding A for 4 days, finding
B for 3 days'' without an explicit order).
 
Formally, we produce an ordered sequence of non-empty time buckets
\[
  \mathcal{B}(r) = (B_1, B_2, \dots, B_K),
\]
where each finding in \(\mathcal{F}(r)\) is assigned to exactly one bucket
\(B_k\), grouping together findings that occurred at the same point in the
patient's clinical course, and index \(k\) orders the buckets from earliest
to latest. This representation is stored as a JSON
list of buckets, which is the format used by both the generator and verifier
below. Models are evaluated solely on temporal structure (ordering and grouping);
optional evidence snippets are not scored.

\subsection{CRAFT: Iterative Generator--Verifier Framework.}

CRAFT operates as a fully automated iterative loop: the generator
proposes a candidate temporal sequence, the verifier scores it against
structural and temporal constraints and returns targeted feedback, and
the loop repeats until the candidate is accepted or a fixed iteration
budget is exhausted. Figure~\ref{fig:overview}(b) and algorithm \ref{alg:craft_algo} illustrate this
process.

\subsubsection{Generator Agent.}
\label{subsec:generator}


At iteration \(i\), the generator LLM implements
\[
  g:\big(\mathcal{F}(r),\, x_r,\, \text{feedback}^{(i-1)}\big)
  \;\longrightarrow\; \hat{\mathcal{B}}^{(i)}(r),
\]
where \(\text{feedback}^{(i-1)}\) is empty at \(i{=}1\) and contains verifier
guidance thereafter.

CRAFT-Full uses a full-regeneration generator: a single
prompt template with full task instructions at each iteration, with verifier
feedback appended to the context when available.
The generator receives: (i) a task description and definition of temporal
progression, including phenomena that should \emph{not} be treated as
progression; (ii) the finding list \(\mathcal{F}(r)\), which must appear exactly as given in the output; and (iii) the free-text narrative \(x_r\). It outputs
a JSON bucket sequence using only findings from \(\mathcal{F}(r)\) and
avoiding unsupported temporal inferences.

\subsubsection{Verifier Agent.}
\label{subsec:verifier}
 
The verifier implements
\[
  v:\big(\mathcal{F}(r),\, x_r,\, \hat{\mathcal{B}}^{(i)}(r)\big)
  \rightarrow \big(\text{decision}^{(i)},\, \text{feedback}^{(i)},\, \text{score}^{(i)}\big),
\]
where \(\text{decision}^{(i)} \in \{\text{ACCEPT}, \text{REVISE}\}\),
\(\text{feedback}^{(i)}\) describes issues to fix, and
\(\text{score}^{(i)} \in \{0,\dots,5\}\) supports early stopping.
The verifier first calls \texttt{FormatTool}, a deterministic helper
module that normalizes the raw generator output into the JSON bucket
schema without altering temporal content, then applies an additive rubric that
assigns one point each for: valid JSON with earliest$\rightarrow$latest ordering; placing
non-mentioned symptoms as \texttt{``none''} in the final group; using each
symptom exactly once; grouping symptoms that occur around the same time; and
ordering groups according to temporal cues in the narrative. When the score
meets threshold \(\theta\), the verifier returns \(\text{ACCEPT}\); otherwise
it returns \(\text{REVISE}\) with targeted feedback for the next iteration.
If no candidate is accepted within \(T_{\max}\) iterations, the loop
terminates and the last candidate is returned. The complete procedure is
given in Algorithm~\ref{alg:craft_algo}.

\section{Experimental Setup.}
We organize our evaluation around three research questions:

\begin{itemize}[leftmargin=*, itemsep=1pt, topsep=2pt]
    \item \textbf{RQ1 Method effectiveness.} Does CRAFT-Full outperform the baselines (PIVOT, GUIDE) across model tiers, and what drives the difference?
    \item \textbf{RQ2 Model capability.} How well do different LLM backbones perform on MedTempo, and is the capability ordering stable across configurations?
    \item \textbf{RQ3 Vaccine discrepancy.} To what extent do performance and error patterns vary across vaccine types, and are model rankings consistent under vaccine-stratified evaluation?
\end{itemize}

The remainder of this section specifies the evaluation data, the model
backbones and their settings, the implementation of each configuration, and
the metrics used to score predicted timelines against the gold standard.

\subsection{Dataset.}

MedTempo contains 5,347 vaccine adverse-event narratives across
three COVID-19 vaccine types, each paired with a provided symptom list.
We evaluate on the 3,166 reports from MedTempo-T with temporal evidence of distinct
symptom progression; the remaining reports contain no temporal progression
and fall outside the primary benchmark.



\subsection{Models and Settings.}

We evaluate four LLMs: \textit{GPT-4.1}~\cite{openai2025gpt41},
\textit{Claude Sonnet 4.5}~\cite{anthropic2025sonnet45},
\textit{MedGemma-27B}~\cite{sellergren2025medgemma}, and
\textit{Llama-3.3-70B}~\cite{meta2024llama33}. Proprietary models are
accessed via their respective APIs with deterministic decoding. Open-weight
models run locally using Hugging Face \texttt{transformers}~\cite{wolf-etal-2020-transformers} with 4-bit
quantization (NF4 via \texttt{bitsandbytes}~\cite{dettmers2023qlora}).
 
We evaluate five settings that systematically vary the generator and
verifier components to isolate the contribution of each.
Our proposed method, CRAFT-Full, pairs the
full-regeneration generator with the additive rubric verifier
(Section~\ref{subsec:generator}--\ref{subsec:verifier}).
Two baselines represent alternative design choices grounded in
established paradigms:
PIVOT pairs the full-regeneration generator with an anchor-based
verifier inspired by the document-creation-time (DCT) anchoring tradition
in temporal extraction~\cite{dct,cheng-miyao-2018-inducing}, where a fixed
reference point serves as a hub for ordering events.
\textbf{GUIDE} pairs an edit-conditioned generator, which applies
targeted local edits to the previous candidate rather than regenerating
from scratch, with the same anchor-based verifier.
CRAFT-G swaps in the edit-conditioned generator while holding the
additive verifier fixed, isolating the generator's contribution;
CRAFT~w/o~V removes the verification loop entirely, running one
generator pass without feedback, isolating the verifier's contribution.
 
\paragraph{Edit-conditioned generator (CRAFT-G, GUIDE).}
Uses a dedicated initialization prompt at \(i{=}1\), switching to a
lightweight editing prompt that conditions on the previous candidate and
verifier feedback at \(i{>}1\).
 
\paragraph{Anchor-based verifier (PIVOT, GUIDE).}
Treats the vaccination date as the dominant time anchor and scores
conservatively, starting from 5 and subtracting one point for clear
violations: anchor-order contradictions, grouping errors (over-merge or
over-split), or overall inconsistency with temporal cues.

\begin{algorithm}[!tp]
\caption{CRAFT: Iterative generator--verifier loop}
\label{alg:craft_algo}
\begin{algorithmic}
\REQUIRE $\mathcal{F}(r)$, narrative $x_r$, generator $G$, verifier $V$,
         $T_{\max}$, threshold $\theta$
\ENSURE  $\hat{\mathcal{B}}(r)$
\STATE \textit{feedback} $\leftarrow \emptyset$
\FOR{$t = 1,\dots,T_{\max}$}
  \STATE $\text{raw} \leftarrow G\!\left(\mathcal{F}(r),\, x_r,\, \textit{feedback}\right)$
  \STATE $\hat{\mathcal{B}}(r) \leftarrow \textsc{FormatTool}(\text{raw})$
  \STATE $\text{score},\; \textit{feedback} \leftarrow V\!\left(\mathcal{F}(r),\, x_r,\, \hat{\mathcal{B}}(r)\right)$
  \IF{$\text{score} \geq \theta$}
    \RETURN $\hat{\mathcal{B}}(r)$
  \ENDIF
\ENDFOR
\RETURN $\hat{\mathcal{B}}(r)$ \COMMENT{return last candidate}
\end{algorithmic}
\end{algorithm}

\subsection{Implementation Details.}
All runs use fixed prompt templates that enforce a strict JSON schema and require every symptom in the provided list to appear exactly once, grouping symptoms into the same stage when the narrative does not support an internal order. We use deterministic decoding (no sampling) with \texttt{max\_new\_tokens}=512. 
\begin{table*}[!tp]
\centering
\scriptsize
\setlength{\tabcolsep}{2.5pt}
\renewcommand{\arraystretch}{1.05}
\caption{General results stratified by vaccine type. Metrics in \%.  \textbf{Bold} = best, \underline{underline} = second best within each model block. $^\dagger$Proposed method.}
\vspace{-2mm}

\label{tab:temporal-round2-vax}
\resizebox{\textwidth}{!}{%
\begin{tabular}{c l  c c c c   c c c  c  c c c  c  c c c}
\toprule
\multirow{2}{*}{\textbf{Model}} & \multirow{2}{*}{\textbf{Setting}}
& \multicolumn{3}{c}{\textbf{Janssen (n=1,164)}}&
& \multicolumn{3}{c}{\textbf{Moderna (n=983)}}&
& \multicolumn{3}{c}{\textbf{Pfizer (n=1,019)}}&
& \multicolumn{3}{c}{\textbf{Total}} \\
\cmidrule(lr){3-5}\cmidrule(lr){7-9}
\cmidrule(lr){11-13}\cmidrule(lr){15-17}
& & \textbf{EM$\uparrow$} & \textbf{LCCS$\uparrow$} & $\boldsymbol{\tau_b}\uparrow$&
  & \textbf{EM$\uparrow$} & \textbf{LCCS$\uparrow$} & $\boldsymbol{\tau_b}\uparrow$&
  & \textbf{EM$\uparrow$} & \textbf{LCCS$\uparrow$} & $\boldsymbol{\tau_b}\uparrow$&
  & \textbf{EM$\uparrow$} & \textbf{LCCS$\uparrow$} & $\boldsymbol{\tau_b}\uparrow$ \\
\midrule
\multirow{5}{*}{\rotatebox[origin=c]{90}{GPT-4.1}}
& PIVOT               & 33.65 & 59.56 & 58.89 & & \underline{38.53} & \underline{60.73} & 55.43 & & \underline{31.89} & 57.44 & \underline{57.00} & & \underline{34.60} & 59.24 & 57.20 \\
& GUIDE               & \underline{33.99} & \underline{61.02} & \textbf{60.07} & & 36.70 & 60.68 & \textbf{56.48} && 31.30 & \underline{57.57} & \textbf{57.85} && 33.97 & \underline{59.80} & \textbf{58.24} \\
& CRAFT$_{\textup{w/o\,V}}$             & 26.49 & 42.94 & 45.79 && 27.93 & 43.51 & 43.65 && 24.51 & 40.06 & 42.37 && 26.30 & 42.19 & 44.02 \\
& CRAFT$_{\textup{G}}$             & 29.68 & 55.71 & 55.92 && 33.94 & 56.52 & 52.08 && 29.92 & 54.86 & 53.31 && 31.08 & 55.69 & 53.88 \\
& \textbf{CRAFT-Full}$^\dagger$ & \textbf{34.43} & \textbf{61.68} & \underline{59.58} && \textbf{40.06} & \textbf{63.84} & \underline{56.45} && \textbf{32.68} & \textbf{58.93} & 56.97 && \textbf{35.61} & \textbf{61.46} & \underline{57.77} \\
\midrule

\multirow{5}{*}{\rotatebox[origin=c]{90}{Llama-70B}}
& PIVOT               & \underline{24.76} & \underline{50.18} & 51.98 && \underline{30.58} & \underline{52.81} & 49.53 && \textbf{26.77} & \underline{50.33} & \underline{52.13} & &\underline{27.22} & \underline{51.04} & 51.27 \\
& GUIDE               & 23.73 & 48.59 & 50.81 && 30.28 & 51.40 & 48.53 && 24.70 & 47.94 & 49.17 && 26.08 & 49.25 & 49.58 \\
& CRAFT$_{\textup{w/o\,V}}$             & 18.64 & 41.34 & \textbf{54.10} && 23.55 & 44.68 & \textbf{51.59} && 20.18 & 41.08 & 51.12 && 20.66 & 42.30 & \underline{52.36} \\
& CRAFT$_{\textup{G}}$             & 23.81 & 48.55 & 50.64 && 30.48 & 51.49 & 48.70 && \underline{24.90} & 47.97 & 49.20 && 26.24 & 49.28 & 49.57 \\
& \textbf{CRAFT-Full}$^\dagger$ & \textbf{25.88} & \textbf{52.06} & \underline{53.47} && \textbf{31.90} & \textbf{53.89} & \underline{50.69} && \textbf{26.77} & \textbf{51.49} & \textbf{52.93} & &\textbf{28.04} & \textbf{52.45} & \textbf{52.43} \\
\midrule

\multirow{5}{*}{\rotatebox[origin=c]{90}{MedGemma}}
& PIVOT               & \underline{17.95} & \underline{39.67} & 43.63 && \underline{25.89} & \underline{43.65} & \textbf{42.08} & &\underline{19.00} & \underline{39.37} & 41.92 & &\underline{20.75} & 40.81 & \textbf{42.60} \\
& GUIDE               & 17.52 & 39.56 & 42.33 && 23.96 & 41.34 & 39.29 && \textbf{19.39} & 38.53 & 40.36 && 20.12 & 39.78 & 40.75 \\
& CRAFT$_{\textup{w/o\,V}}$             & 13.03 & 33.21 & \textbf{44.52} && 15.90 & 36.38 & 40.61 && 14.67 & 35.00 & \textbf{42.20} && 14.44 & \textbf{44.36} & 42.56 \\
& CRAFT$_{\textup{G}}$             & 15.53 & 35.90 & 38.54 && 21.92 & 37.83 & 35.61 && 18.11 & 35.73 & 36.68 && 18.35 & 36.44 & 37.03 \\
& \textbf{CRAFT-Full}$^\dagger$ & \textbf{18.29} & \textbf{40.20} & \underline{43.84} & & \textbf{25.99} & \textbf{43.91} & \underline{41.59} && 18.80 & \textbf{39.49} & \underline{42.09}&& \textbf{20.85} & \underline{41.13} & \underline{42.57} \\
\midrule

\multirow{5}{*}{\rotatebox[origin=c]{90}{Claude-4.5}}
& PIVOT               & 34.51 & 57.34 & 52.71 && 40.37 & 59.47 & 51.03 && \underline{34.84} & 55.58 & 51.66 & & 36.44 & 57.44 & 51.85 \\
& GUIDE               & 34.69 & 54.98 & 50.53 && 38.23 & 56.61 & 48.03& & 34.35 & 53.09 & 48.21 && 35.68 & 54.88 & 49.00 \\
& CRAFT$_{\textup{w/o\,V}}$             & \textbf{37.19} & 57.73 & 56.03 && \underline{40.88} & 59.85 & \underline{54.20} && \textbf{35.63} & 55.23 & \underline{53.06} && \textbf{37.83} & 57.58 & 54.51 \\
& CRAFT$_{\textup{G}}$             & 33.74 & \underline{61.04} & \textbf{57.58} && 40.67 & \underline{63.39} & 53.99 && 34.06 & \textbf{59.57} & \textbf{54.50} && 35.99 & \underline{61.30} & \textbf{55.47} \\
& \textbf{CRAFT-Full}$^\dagger$ & \underline{35.89} & \textbf{61.82} & \underline{56.68} && \textbf{41.18} & \textbf{64.51} & \textbf{54.98} && 34.65 & \underline{58.54} & 52.92 && \underline{37.14} & \textbf{61.60} & \underline{54.94} \\
\bottomrule
\end{tabular}%
}
\vspace{-3mm}
\end{table*}

For generator--verifier configurations, we run up to \texttt{max\_iter}=4 refinement iterations and accept outputs when the verifier score is at least \textbf{\(\theta=3\)}; the parameter-selection procedure for \texttt{max\_iter} and \(\theta\) is summarized in Appendix \ref{sec:appendix-hyperparams}. Full prompt templates for CRAFT-Full are provided in Appendix \ref{app:prompts}. More implementation details can be found at \url{https://github.com/LEAF-Lab-Stevens/TemporalAnalysis}. 

Open-weight experiments are run on a workstation equipped with two NVIDIA RTX A5000 GPUs (24GB each). Large models are loaded with device sharding across both GPUs and 4-bit quantization. API-based experiments are executed using the official Python SDKs for the corresponding providers.

\subsection{Evaluation Metrics.}
 
We evaluate temporal sequences as ordered lists of buckets, where each bucket is a set of items and within-bucket order is trivial. Let the gold sequence be \(G=(G_1,\dots,G_m)\) and the prediction \(P=(P_1,\dots,P_n)\). After normalization \(\phi(\cdot)\) (e.g., lower-casing), write \(\tilde{G}_i=\{\phi(x)\mid x\in G_i\}\) and \(\tilde{P}_j=\{\phi(x)\mid x\in P_j\}\); let \(r_G(x)\) and \(r_P(x)\) denote the bucket rank of item \(x\) in gold and prediction, and \(S\) the set of items appearing in both. Within-bucket duplicates are ignored; cross-bucket duplicates are resolved by a check-and-fix module \texttt{FormatTool}.

\smallskip
\textbf{Strict Exact Match (EM)~\cite{rajpurkar2016squad}.}\par

Pass/fail: the prediction must exactly reproduce the gold segmentation and inter-bucket order (within-bucket order ignored).

\[
\mathrm{EM}(G,P)=
\begin{cases}
1, &m=n\ \text{and}\ \forall\,i\in\{1,...,m\}: \tilde{G}_i=\tilde{P}_i,\\
0, & \text{otherwise.}
\end{cases}
\]

\smallskip
\textbf{Kendall's \(\tau_b\)~\cite{kendall1945treatment}.}\par

For each unordered pair \(\{x,y\}\subset S\), the pair is \emph{concordant} if \(\operatorname{sign}(r_G(y){-}r_G(x))=\operatorname{sign}(r_P(y){-}r_P(x))\neq 0\), and \emph{discordant} if the signs are non-zero and opposite. Let \(N_C\), \(N_D\) be the concordant and discordant counts, and \(T_G\), \(T_P\) the pair counts tied in gold and prediction:
\[
\tau_b=\frac{N_C-N_D}{\sqrt{(N_C+N_D+T_G)\,(N_C+N_D+T_P)}}\in[-1,1].
\]
Ranges from \(-1\) (complete reversal) to \(+1\) (perfect agreement); ties from within-bucket equivalence are handled explicitly.

\smallskip
\textbf{Group-Aware LCCS~\cite{gusfield1997algorithms}.}\par
Treating each bucket as a token, LCCS is the longest common contiguous subsequence of \((\tilde{G}_1,\dots,\tilde{G}_m)\) and \((\tilde{P}_1,\dots,\tilde{P}_n)\):
\[
L=\max\bigl\{\,\ell\mid\exists\,i,j:\,(\tilde{G}_i,...,\tilde{G}_{i+\ell-1})=(\tilde{P}_j,...,\tilde{P}_{j+\ell-1})\bigr\}.
\]
This metric complements \(\tau_b\) by rewarding \textbf{unbroken} spans of perfectly matched phases, penalising isolated segmentation errors.
\section{Results.}
This section reports our empirical findings, organized around the three
research questions. We first compare CRAFT-Full against the baselines, then
examine how the four backbones compare to one another, and then test whether
these patterns hold when performance is broken out by vaccine type. Two
further subsections follow: an ablation that isolates the contributions of the
verifier and the generator, and case studies that illustrate the behavior
behind the aggregate numbers.

\subsection{RQ1: CRAFT-Full vs.\ Baselines.}
\label{sec:rq1}

Table~\ref{tab:temporal-round2-vax} reports final performance across all four

\begin{table*}[!t]
\centering
\scriptsize
\setlength{\tabcolsep}{3.2pt}
\renewcommand{\arraystretch}{1.05}
\caption{Results at each iteration per model and setting.
$M@t$: metric $M\in\{\text{EM},\text{LCCS},\tau_b\}$ when the
generate--verify loop is capped at $t$ iterations. AvgIters: mean iterations executed under
$t{=}4$. Metrics in \%. \underline{Underline}: best per setting
across iterations. \textbf{\underline{Bold underline}}: best per
model per metric. $^\dagger$Proposed method.}

\vspace{-2mm}

\label{tab:ablation-allmetrics-trend}
\resizebox{\textwidth}{!}{%
\begin{tabular}{c l c 
                c c c c   
                c c c c   
                c c c c}   
\toprule
\textbf{LLM} & \textbf{Setting} & \textbf{AvgIters}
& \textbf{EM@1} & \textbf{EM@2} & \textbf{EM@3} & \textbf{EM@4}
& \textbf{LCCS@1} & \textbf{LCCS@2} & \textbf{LCCS@3} & \textbf{LCCS@4}
& $\boldsymbol{\tau_b}@1$ & $\boldsymbol{\tau_b}@2$ & $\boldsymbol{\tau_b}@3$ & $\boldsymbol{\tau_b}@4$ \\
\midrule
\multirow{5}{*}{\rotatebox[origin=c]{90}{GPT-4.1}}
& PIVOT & 1.2734
& 29.72 & \underline{34.73} & \underline{34.73} & 34.60
& 52.19 & \underline{59.44} & 59.31 & 59.24
& 50.74 & \underline{57.29} & 57.28 & 57.20 \\
& GUIDE & 1.2348
& \underline{34.92} & 34.22 & 34.06 & 33.97
& \underline{60.96} & 60.18 & 59.97 & 59.80
& \textbf{\underline{58.91}} & 58.44 & 58.36 & 58.24 \\
& CRAFT$_{\textup{w/o\,V}}$ & 1.0000
& \underline{26.30} & --- & --- & ---
& \underline{42.19} & --- & --- & ---
& \underline{44.02} & --- & --- & --- \\
& CRAFT$_{\textup{G}}$ & 2.9407
& \underline{34.89} & 32.79 & 31.78 & 31.08
& \underline{60.90} & 58.16 & 56.61 & 55.69
& \underline{58.33} & 55.87 & 54.40 & 53.88 \\
& \textbf{CRAFT-Full}$^\dagger$ & 2.9864
& 26.90 & 34.89 & \textbf{\underline{35.61}} & \textbf{\underline{35.61}}
& 47.10 & 60.94 & 61.45 & \textbf{\underline{61.46}}
& 45.71 & \underline{58.19} & 58.16 & 57.77 \\
\midrule

\multirow{5}{*}{\rotatebox[origin=c]{90}{Llama-70B}}
& PIVOT & 1.2022
& 26.90 & \underline{27.22} & \underline{27.22} & \underline{27.22}
& 50.13 & 51.00 & \underline{51.04} & \underline{51.04}
& 50.07 & 51.21 & \underline{51.27} & \underline{51.27} \\
& GUIDE & 1.2148
& \underline{26.08} & 26.08 & 26.08 & 26.08
& 49.23 & 49.25 & 49.25 & \underline{49.28}
& 49.54 & \underline{49.59} & 49.58 & 49.58 \\
& CRAFT$_{\textup{w/o\,V}}$ & 1.0000
& \underline{20.66} & --- & --- & ---
& \underline{42.30} & --- & --- & ---
& \underline{52.36} & --- & --- & --- \\
& CRAFT$_{\textup{G}}$ & 1.2643
& 26.08 & \underline{26.27} & 26.24 & 26.24
& 49.23 & \underline{49.30} & 49.26 & 49.28
& 49.54 & 49.52 & 49.53 & \underline{49.57} \\
& \textbf{CRAFT-Full}$^\dagger$ & 1.2136
& 27.47 & 27.95 & 28.01 & \textbf{\underline{28.04}}
& 50.92 & 52.32 & 52.42 & \textbf{\underline{52.45}}
& 50.28 & 52.37 & 52.41 & \textbf{\underline{52.43}} \\
\midrule
 
\multirow{5}{*}{\rotatebox[origin=c]{90}{MedGemma}}
& PIVOT & 1.1496
& 20.66 & \underline{20.75} & \underline{20.75} & \underline{20.75}
& 40.56 & 40.82 & \underline{40.83} & 40.81
& 42.08 & 42.60 & \textbf{\underline{42.63}} & 42.60 \\
& GUIDE & 1.1179
& \underline{20.15} & 20.12 & 20.12 & 20.12
& \underline{39.80} & 39.78 & 39.78 & 39.78
& \underline{40.86} & 40.76 & 40.75 & 40.75 \\
& CRAFT$_{\textup{w/o\,V}}$ & 1.0000
& \underline{14.44} & --- & --- & ---
& \textbf{\underline{44.36}} & --- & --- & ---
& \underline{42.56} & --- & --- & --- \\
& CRAFT$_{\textup{G}}$ & 1.5542
& \underline{20.15} & 18.47 & 18.35 & 18.35
& \underline{39.80} & 36.49 & 36.44 & 36.44
& \underline{40.86} & 37.13 & 37.03 & 37.03 \\
& \textbf{CRAFT-Full}$^\dagger$ & 1.5032
& 20.66 & 20.47 & \textbf{\underline{20.85}} & \textbf{\underline{20.85}}
& 40.56 & 40.64 & \underline{41.13} & \underline{41.13}
& 42.08 & 42.32 & \underline{42.57} & \underline{42.57} \\
\midrule

\multirow{5}{*}{\rotatebox[origin=c]{90}{Claude-4.5}}
& PIVOT & 1.9861
& \underline{37.58} & 35.33 & 36.66 & 36.44
& \underline{62.65} & 56.92 & 57.81 & 57.44
& \underline{56.73} & 52.61 & 52.55 & 51.85 \\
& GUIDE & 1.9943
& \textbf{\underline{40.37}} & 35.17 & 36.47 & 35.68
& \textbf{\underline{65.37}} & 55.56 & 55.75 & 54.88
& \textbf{\underline{58.57}} & 51.42 & 49.97 & 49.00 \\
& CRAFT$_{\textup{w/o\,V}}$ & 1.0000
& \underline{37.83} & --- & --- & ---
& \underline{57.58} & --- & --- & ---
& \underline{54.51} & --- & --- & --- \\
& CRAFT$_{\textup{G}}$ & 3.6350
& \underline{40.08} & 35.90 & 37.48 & 35.99
& \underline{65.35} & 61.63 & 62.51 & 61.30
& \underline{58.40} & 55.59 & 56.07 & 55.47 \\
& \textbf{CRAFT-Full}$^\dagger$ & 3.4623
& 36.57 & 36.57 & \underline{37.42} & 37.14
& 61.32 & 61.49 & \underline{61.65} & 61.60
& \underline{55.74} & 54.83 & 55.05 & 54.94 \\

\bottomrule
\end{tabular}%
}
\vspace{-3mm}
\end{table*}

models and five settings. Against PIVOT, CRAFT-Full gains $+$1.0 EM points for GPT-4.1 (35.61 vs.\ 34.60), $+$0.8
for Llama-3.3-70B (28.04 vs.\ 27.22), $+$0.7 for Claude Sonnet~4.5 (37.14
vs.\ 36.44), and $+$0.1 for MedGemma-27B (20.85 vs.\ 20.75). Margins over
GUIDE are uniformly larger (GPT-4.1: $+$1.6; Llama: $+$2.0; Claude: $+$1.5;
MedGemma: $+$0.7). CRAFT-Full achieves the highest EM in every model block compared to baselines, confirming it
as the strongest configuration. Although baselines such as PIVOT and
GUIDE occasionally match or exceed CRAFT-Full on $\tau_b$ or LCCS, these
metrics credit partial ordering agreement and can score highly even when
the full trajectory structure is incorrect. EM, which requires the
entire stage-wise grouping and ordering to match gold, is the most
demanding metric for this task and the one on which CRAFT-Full
consistently leads.

Table~\ref{tab:ablation-allmetrics-trend} explains how CRAFT-Full wins.
CRAFT-Full actively uses its refinement budget: AvgIters reaches 2.99 for
GPT-4.1 and 3.46 for Claude, and performance builds 
across iterations. For GPT-4.1, EM rises from 26.90 at $i{=}1$ to 35.61 at
$i{=}4$, a gain of 8.7 points across rounds. In contrast, PIVOT and GUIDE converge after a single pass in most instances (AvgIters $\approx$ 1.2--2.0)
because the anchor-based verifier accepts outputs quickly without substantive
ordering improvement. For GPT-4.1, PIVOT peaks at EM@2$=$34.73 and GUIDE
degrades from its peak of 34.92 at $i{=}1$. This confirms that CRAFT-Full's
additive rubric provides richer, more actionable feedback that sustains
improvement across the full refinement budget, whereas the anchor-based
verifier's narrower signal cannot drive continued gains beyond the first pass.

\begin{figure}[!t]
    \centering
    \begin{subfigure}[t]{0.48\columnwidth}
        \centering
        \includegraphics[width=\linewidth]{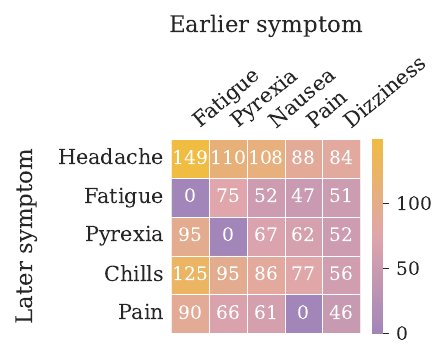}
        \vspace{-7mm}
        \caption{Ground Truth}
    \end{subfigure}\hfill
    \begin{subfigure}[t]{0.48\columnwidth}
        \centering
        \includegraphics[width=\linewidth]{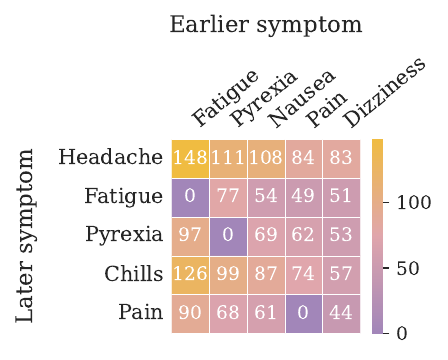}
        \vspace{-7mm}

        \caption{Claude, CRAFT-Full}
    \end{subfigure}
    \vspace{-2mm}

    \caption{Before--after symptom-transition frequencies of each later symptom (columns) given the first
    symptom of the report (rows; restricted to the most frequent first
    symptoms in the dataset): (a)~ground truth; (b)~CRAFT-Full on Claude}
    \label{fig:heatmap_overall}
    \vspace{-5mm}
\end{figure}

Figure~\ref{fig:heatmap_overall} shows that the gold and CRAFT-Full
transition matrices share a similarly diffuse distribution across
symptom pairs and differ only in small cell values, confirming that
CRAFT-Full preserves progression patterns at the population scale.
Since this aggregate view does not reveal where refinement changes
predictions, we turn to an instance level case study in Section~\ref{sec:case_study}.

\subsection{RQ2: Overall Model Performance.}
\label{sec:rq2}
 
The capability ordering is stable across all configurations and metrics:
Claude Sonnet~4.5 $>$ GPT-4.1 $>$ Llama-3.3-70B $>$ MedGemma-27B.
Under CRAFT-Full, Claude achieves the highest total EM (37.14), followed by
GPT-4.1 (35.61), Llama (28.04), and MedGemma (20.85). This ordering holds
without exception across all three metrics and all vaccine strata, indicating
that MedTempo reliably differentiates model tiers.


Beyond the stable ranking, models differ notably in how they respond to
iterative refinement. GPT-4.1 benefits most from additional iterations:
under CRAFT-Full, EM climbs from 26.90 at $i{=}1$ to 35.61 at $i{=}4$
($+$8.7 points), with LCCS and $\tau_b$ following similar upward trends
(47.10$\to$61.46 and 45.71$\to$57.77 respectively). In contrast, Claude
starts strong (EM 36.57 at $i{=}1$) but gains only $+$0.6 points across
iterations, suggesting its first-pass outputs already capture most of the
temporal structure. Llama and MedGemma show similarly flat iteration
curves (EM gains of $+$0.6 and $+$0.2 respectively), but for a different
reason: both converge quickly (AvgIters 1.2--1.5), indicating that the
verifier accepts their outputs early rather than driving further
improvement. This divergence between models that saturate from high
initial quality (Claude) and those that stall from limited capacity to
act on feedback (Llama, MedGemma) highlights that iteration utility is
tied to model capability.

\begin{figure}[!t]
    \centering
    \includegraphics[width=0.5\textwidth]{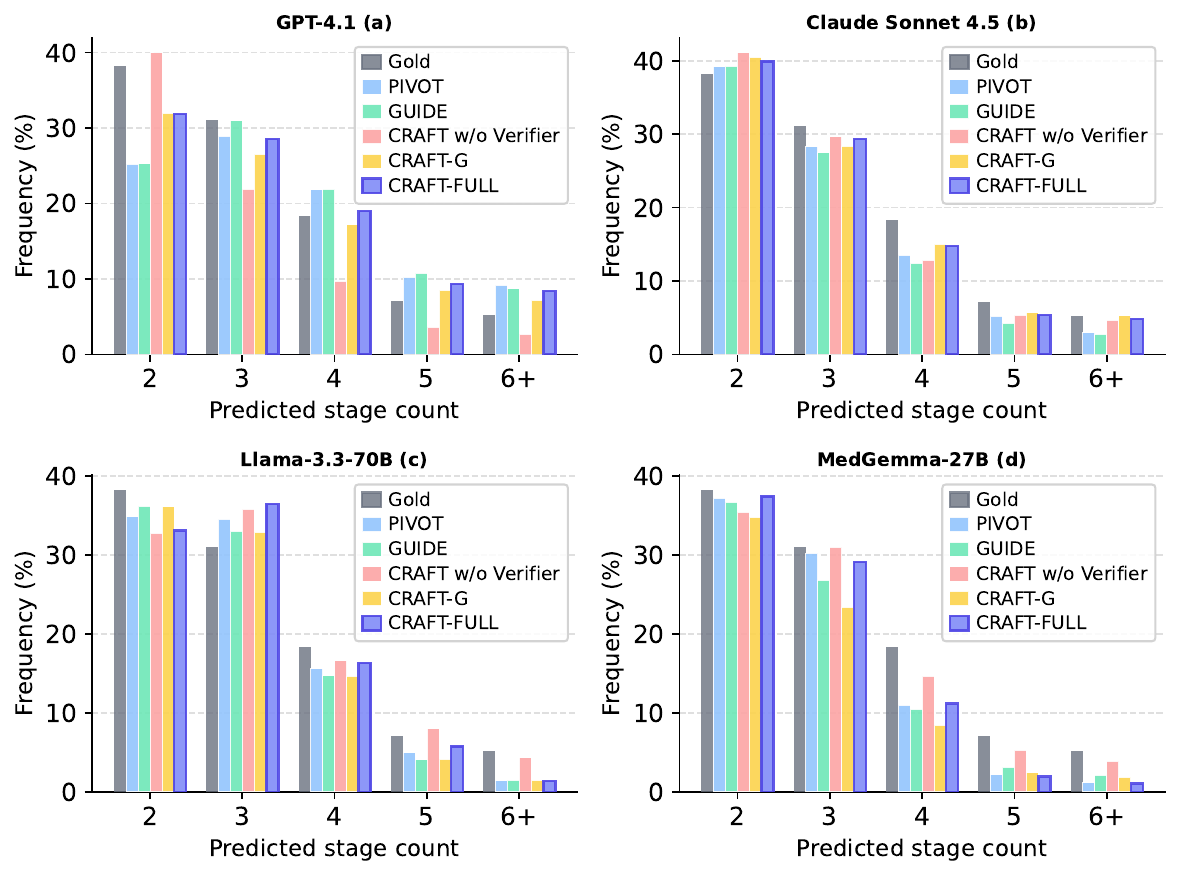}
    \vspace{-5mm}

    \caption{Predicted stage-count distributions (frequency \%) under all five
settings vs.\ gold, per model. All models under-segment relative to gold;
the bias is more pronounced for weaker models (MedGemma-27B, Llama-3.3-70B)
than for stronger ones (GPT-4.1, Claude Sonnet~4.5).}
    \label{fig:stage}
    \vspace{-5mm}

\end{figure}

Figure~\ref{fig:stage} shows that distributional fidelity to gold tracks
the capability ordering: Claude overlaps gold across all
settings, while Llama and MedGemma fall short of gold at
stage counts of four and above regardless of verifier choice. GPT-4.1
is the one tier where verifier choice visibly reshapes the distribution,
with CRAFT~w/o~V peaking at stage~2 and CRAFT-Full recovering a spread
close to gold.




 \subsection{RQ3: Vaccine-Stratified Performance.}
\label{sec:rq3}
 
Performance is consistently stratified by vaccine type across all models and
configurations: Moderna narratives yield the highest EM in every case, followed
by Janssen, with Pfizer-BioNTech systematically lowest. Under CRAFT-Full, the
Moderna--Pfizer gap is 7.4 points for GPT-4.1 (40.06 vs.\ 32.68), 6.5 for
Claude Sonnet~4.5 (41.18 vs.\ 34.65), 5.1 for Llama-3.3-70B (31.90 vs.\
26.77), and 7.2 for MedGemma-27B (25.99 vs.\ 18.80). This ordering is
preserved across all settings and all four models without exception, indicating
a systematic corpus-level source of difficulty rather than a model-specific
effect.

The vaccine gap is more pronounced on EM than on LCCS and $\tau_b$, suggesting
that models make segmentation errors on Pfizer narratives specifically rather
than systematically misranking symptoms within groups. Since EM requires an
exact match on both bucket composition and inter-bucket order while LCCS rewards
contiguous correct spans and $\tau_b$ captures global pairwise ordering, the
metric divergence points to Pfizer narratives being harder to segment into
correct temporal groups rather than harder to order within those groups. The
capability ordering Claude $>$ GPT-4.1 $>$ Llama $>$ MedGemma holds for every
vaccine type under every configuration.

\subsection{Ablation Study.}
\label{sec:ablation}

Using Table~\ref{tab:ablation-allmetrics-trend}, we analyze two ablation groups: the verifier ablation (CRAFT-Full vs.\ CRAFT~w/o~V), which holds the generator fixed, and the generator ablation (CRAFT-G vs.\ CRAFT-Full), which holds the verifier fixed.
 
\paragraph{Verifier contribution.}
Removing the verification loop causes substantial performance drops for three
of four models. CRAFT-Full gains $+$9.3 EM over CRAFT~w/o~V for GPT-4.1
(35.61 vs.\ 26.30), $+$7.4 for Llama-3.3-70B (28.04 vs.\ 20.66), and $+$6.4
for MedGemma-27B (20.85 vs.\ 14.44). For Llama and MedGemma, virtually all
gain is captured at $i{=}1$: the verifier's

\begin{table}[!tp]
\centering
\small
\setlength{\tabcolsep}{4pt}
\renewcommand{\arraystretch}{1.35}
\caption{Iteration trace for Example~1 (GPT-4.1, CRAFT-Full).
$\checkmark$~=~matches gold; $\times$~=~grouping error.}
    \vspace{-2mm}

\label{tab:casestudy1}
\begin{tabular}{c c p{5.6cm}}
\toprule
\textbf{Iter} & \textbf{Score} & \textbf{Predicted Stages} \\
\midrule
Gold & --- &
  \{Insomnia\} $\to$ \{Muscle disorder\}
  $\to$ \{Herpes zoster\} $\to$ \{Memory impairment\} \\
\midrule
1 & 2/5 &
  \{Insomnia\} $\to$ \{Muscle disorder\}
  $\to$ \{\textbf{Herpes zoster, Memory impairment}\}
  \textcolor{red}{$\times$} \\
\midrule
2 & 3/5 &
  \{Insomnia\} $\to$ \{Muscle disorder\}
  $\to$ \{Herpes zoster\} $\to$ \{Memory impairment\}
  \textcolor{green!60!black}{$\checkmark$} \\
\bottomrule
\end{tabular}


\end{table}
 
first-pass feedback corrects schema violations in unverified outputs, after
which the output is accepted without substantive ordering improvement.
For GPT-4.1, gains accumulate steadily through $i{=}3$ (AvgIters~=~2.99),
confirming that the verification loop provides genuine multi-iteration value
for capable models. For Claude Sonnet~4.5, CRAFT-Full (37.14) falls slightly
below CRAFT~w/o~V (37.83): the fixed threshold $\theta{=}3$ does not recognize
Claude's near-correct first-pass output as satisfactory, forcing continued
revision that introduces errors rather than correcting them. This highlights 
the importance of threshold calibration for further improvement.
 
\paragraph{Generator contribution.}
Replacing the full-regeneration generator with the edit-conditioned variant
(CRAFT-G) consistently degrades final performance. For GPT-4.1, CRAFT-G
achieves a strong EM of 34.89 at $i{=}1$ but degrades monotonically to 31.08
by $i{=}4$ ($-$3.8 points), while CRAFT-Full builds to 35.61. The
edit-conditioned prompt is effective for initialisation but too constrained to
sustain quality under continued verifier feedback across the full budget.
For MedGemma-27B, CRAFT-G is actively harmful: EM drops from 20.15 at
$i{=}1$ to 18.35 by $i{=}2$, indicating the model misinterprets edit-only
instructions and degrades its own output under revision. CRAFT-Full avoids
this across all tiers by regenerating from full task context at every
iteration.

\subsection{Case Study.}
\label{sec:case_study}
The ablation analysis above quantifies verifier and generator contributions
across models and iterations at the aggregate level, but does not reveal how
the feedback loop operates on individual instances. To provide insight into
the refinement mechanism, we present a representative example showing how
verifier feedback guides the generator toward the correct trajectory across
iterations. An additional case study is provided in the Appendix \ref{appendix:case2}.
 
\smallskip
\noindent\textit{Example (GPT-4.1, CRAFT-Full):}\par


\noindent\textbf{Symptom list:} Insomnia, Muscle disorder, Herpes zoster,
Memory impairment

\smallskip
\noindent\textbf{Narrative (abridged):} \textit{``On 20Feb2021, the patient
experienced could not sleep. In Feb2021, the patient experienced effected
upper part of body, arms, neck, shoulders, severe muscle condition, had tiny
bit of shingles on my side, did not have memory of anything that happened /
was out of my mind [\ldots] They took her to the hospital and then the patient
underwent MRI to see if her brain was okay.''}

\smallskip
\noindent\textbf{Iteration 1} \hfill \textsc{Score: 2/5 --- below
$\theta$, continue}

\noindent\textbf{Verifier feedback:} \textit{``Symptoms are grouped in
temporal order and all are present, but `Herpes zoster' and `Memory
impairment' should be in separate groups as they are not clearly described
as occurring at the same time [\ldots] For full marks, ensure symptoms
grouped together are clearly simultaneous per the text.''}

\smallskip
\noindent\textbf{Iteration 2} \hfill \textsc{Score: 3/5 ---
$s \geq \theta$, early stop \checkmark}

\noindent\textbf{Verifier feedback:} ---

\smallskip
\noindent At $i{=}1$, the model correctly orders all four symptom groups
but over-merges Herpes zoster and Memory impairment into a single stage,
receiving a score of 2. The score of 2 rather than 3 reflects that the
over-merge constitutes two rubric violations simultaneously: grouping
symptoms without clear simultaneous support, and failing to maintain
strict earliest-to-latest ordering within the merged group. CRAFT-Full's
verifier precisely identifies the grouping error and provides a targeted
one-sentence correction, demonstrating that its additive rubric produces
informative, actionable feedback even when the overall ordering direction
is already correct. At $i{=}2$ the model splits the two symptoms into
separate stages, exactly matching the gold standard and receiving a score
of 3, which meets the acceptance threshold $\theta{=}3$ and triggers early
stopping. This example illustrates that a single round of targeted feedback
is sufficient for a capable model to correct a local grouping error within
the refinement budget. Table~\ref{tab:casestudy1} summarizes the
predicted stages across iterations.

\section{Conclusion.}


We presented CRAFT, a generator--verifier LLM framework for iterative temporal reasoning over clinical narratives, and MedTempo, an expert-annotated benchmark of 5,347 adverse-event narratives, to evaluate structured symptom trajectory construction under sparse temporal anchoring. Experiments across four LLM backbones show that CRAFT consistently improves temporal ordering accuracy. Ablation analysis confirms that both generator and verifier components contribute meaningfully, with CRAFT-Full emerging as the strongest configuration across all model tiers. Our results also reveal that refinement behavior varies by model capability, suggesting that adaptive verification strategies may further improve performance. Future work will leverage the 2,181 non-temporal reports from MedTempo-NT as supervision for a learned temporal-evidence identification module, extending CRAFT into a unified end-to-end framework for automatic progression detection and stage-wise temporal ordering.

\section*{Acknowledgments.}
This work was supported in part by the US National Science Foundation grant  IIS-2245907, IIS-2047843, IIS-2437621, and an Amazon Research Award, Fall 2024.

\newpage
\bibliographystyle{siamplain}
\bibliography{example_references}

\newpage
\appendix

\section{Appendix.}\label{sec:appendix}

\subsection{Data Sampling.}\label{appendix:data_sampling}

Table~\ref{tab:excluded_examples} presents
representative examples of reports removed under each criterion.

\subsection{Hyperparameter Selection.}\label{sec:appendix-hyperparams}

Hyperparameter selection for $T_{\max}$ and $\theta$ was conducted using
GPT-4.1 on a development sample of 100 instances drawn from MedTempo-T.
Due to computational budget constraints, as each sweep configuration
requires multiple LLM calls per instance across both generator and
verifier, we limited the search to a single frontier model rather than
conducting independent sweeps for all four backbones, and to a
representative subset rather than the full evaluation set. The selected
values ($T_{\max}{=}4$, $\theta{=}3$) are applied uniformly across all
models and configurations in the main experiments. While model-specific
tuning could yield marginal gains for individual backbones, uniform
hyperparameters ensure a controlled comparison and reflect a realistic
deployment scenario where per-model calibration may not be feasible.
Tables~\ref{tab:imax_sweep} and~\ref{tab:theta_sweep} report the
development set EM across the swept ranges.

\begin{table*}[!t]
\centering
\caption{Examples of records removed from the dataset. This table presents
representative samples of clinical reports excluded from the final dataset
due to criteria such as brevity, lack of temporal information, or limited
symptom mentions.}
\label{tab:excluded_examples}
\resizebox{\textwidth}{!}{
\begin{tabular}{p{4cm} p{6.2cm} p{7.3cm}}
\toprule
\bf Category & \bf Standard Symptoms & \bf Clinical Text \\
\midrule
\multirow{3}{=}{\bf Reports with 3 or less symptoms}
& [Pharyngeal swelling] & patient called back the next day and stated her throat was swelling and had to take Benadryl. \\
\cmidrule(l){2-3}
& [Dysphagia, Epiglottitis] & Right side of epiglottis swelled up and hinder swallowing pictures taken Benadryl Tylenol taken \\
\cmidrule(l){2-3}
& [Dizziness, Fatigue, Mobility decreased] & extreme fatigue, dizziness,. could not lift my left arm for 72 hours \\
\midrule
\multirow{3}{=}{\bf Reports with less than 10 words}
& [Erythema, Pruritus, Rash, Swelling] & redness, bumps, itchiness, and local swelling \\
\cmidrule(l){2-3}
& [Asthenia, Chills, Headache, Myalgia] & Headache, chills, muscle aches and weakness \\
\cmidrule(l){2-3}
& [Chills, Dizziness, Injection site pain, Myalgia, Pyrexia] & Dizziness, chills, fever, muscle aches, pain at the injection site \\
\midrule
\multirow{3}{=}{\bf Reports with no temporal cues}
& [Headache, Nausea, Pain, Pyrexia, Urticaria, Vomiting] & Fever, headache, body aches, nausea and vomiting. Hives \\
\cmidrule(l){2-3}
& [Dizziness, Headache, Hypoaesthesia, Injection site pain] & Headache, sore arm to injection site, dizzy, numbness to right foot. No medications taken. Continues to have symptoms. \\
\cmidrule(l){2-3}
& [Arthralgia, Chills, Fatigue, Headache, Myalgia, Nausea, Pyrexia, Vomiting] & Repeated shaking with chills, headache, nausea and vomiting, muscle/joint aches, fatigue, fever. \\
\bottomrule
\end{tabular}
}
\end{table*}

\begin{figure*}[!t]
    \centering
    \begin{subfigure}[t]{0.23\textwidth}
        \centering
        \includegraphics[width=\linewidth]{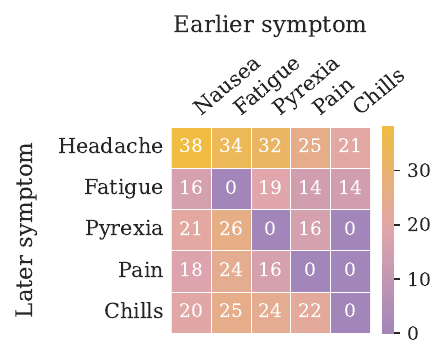}
        \caption*{Pfizer-BioNTech}
    \end{subfigure}\hfill
    \begin{subfigure}[t]{0.23\textwidth}
        \centering
        \includegraphics[width=\linewidth]{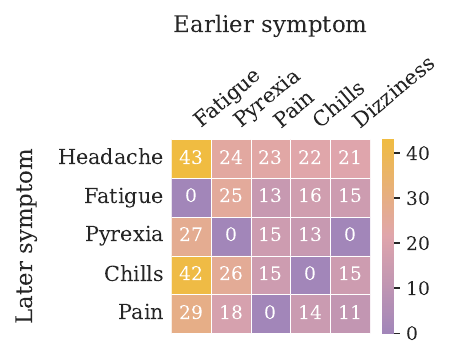}
        \caption*{Moderna}
    \end{subfigure}\hfill
    \begin{subfigure}[t]{0.23\textwidth}
        \centering
        \includegraphics[width=\linewidth]{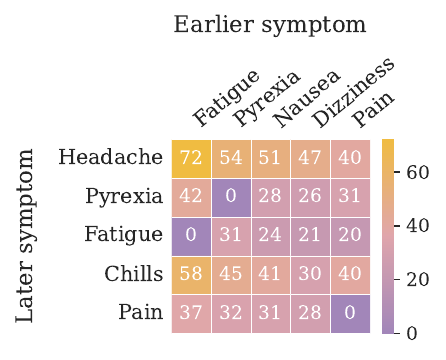}
        \caption*{Janssen}
    \end{subfigure}\hfill
    \begin{subfigure}[t]{0.23\textwidth}
        \centering
        \includegraphics[width=\linewidth]{Figures/Dataset/gold_symptom_heatmap_Overall.pdf}
        \caption*{Overall}
    \end{subfigure}

    \vspace{0.5em}
    \textbf{(a) Ground Truth} \\[0.5em]

    \begin{subfigure}[t]{0.23\textwidth}
        \centering
        \includegraphics[width=\linewidth]{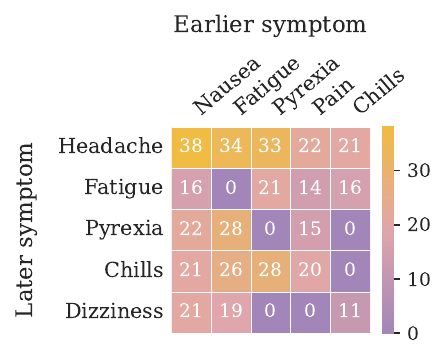}
        \caption*{Pfizer-BioNTech}
    \end{subfigure}\hfill
    \begin{subfigure}[t]{0.23\textwidth}
        \centering
        \includegraphics[width=\linewidth]{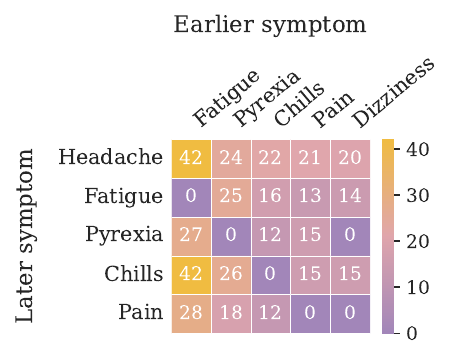}
        \caption*{Moderna}
    \end{subfigure}\hfill
    \begin{subfigure}[t]{0.23\textwidth}
        \centering
        \includegraphics[width=\linewidth]{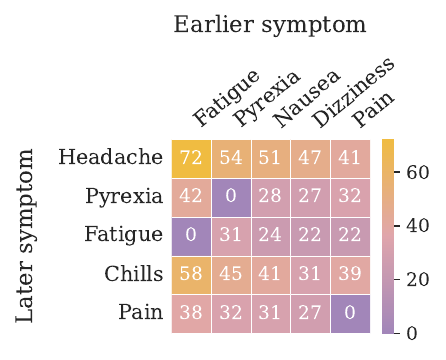}
        \caption*{Janssen}
    \end{subfigure}\hfill
    \begin{subfigure}[t]{0.23\textwidth}
        \centering
        \includegraphics[width=\linewidth]{Figures/Dataset/claude_heatmap/claude_craft_symptom_heatmap_Overall.pdf}
        \caption*{Overall}
    \end{subfigure}

    \vspace{0.5em}
    \textbf{(b) Best Model (Claude Sonnet 4.5, CRAFT-Full)}

    \caption{Before--after symptom relationship heatmaps across vaccine
    types; rows denote the most frequent earlier symptoms and columns
    denote the most frequent subsequent symptoms. Row~(a)
    ground truth; row~(b)CRAFT-Full on Claude}
    \label{fig:before_after_comparison}
\end{figure*}

\begin{table}[H]
\centering
\scriptsize
\setlength{\tabcolsep}{4.0pt}
\renewcommand{\arraystretch}{1.15}
\caption{Development EM (\%) across maximum iteration budget
$T_{\max} \in \{2,\dots,10\}$ for CRAFT. $T_{\max}$ is the upper
cap on generator--verifier loop iterations per instance with early
stopping enabled. GPT-4.1, sample size 100.}
\label{tab:imax_sweep}
\begin{tabular}{lccccccccc}
\toprule
& \textbf{2} & \textbf{3} & \textbf{4} & \textbf{5} &
\textbf{6} & \textbf{7} & \textbf{8} & \textbf{9} & \textbf{10} \\
\midrule
CRAFT
& 12.5 & 20.0 & \textbf{22.5} & 20.0 & 11.0 & 19.5 & 18.5 & 20.0 & 19.0 \\
\bottomrule
\end{tabular}
\vspace{0.8mm}
\begin{minipage}{0.97\linewidth}
\footnotesize
\textbf{Note.} Columns denote $T_{\max}$. Bold indicates the best EM.
$T_{\max}{=}4$ is adopted in all main experiments.
\end{minipage}
\end{table}

\begin{table}[H]
\centering
\scriptsize
\setlength{\tabcolsep}{4.0pt}
\renewcommand{\arraystretch}{1.15}
\caption{Development EM (\%) across verifier acceptance threshold
$\theta \in \{1,2,3,4\}$ for CRAFT. $\theta$ is the minimum verifier
score (0--5 scale) required to accept a candidate output. GPT-4.1,
sample size 100.}
\label{tab:theta_sweep}
\begin{tabular}{l c c c c}
\toprule
& $\boldsymbol{\theta{=}1}$ &
$\boldsymbol{\theta{=}2}$ & $\boldsymbol{\theta{=}3}$ &
$\boldsymbol{\theta{=}4}$ \\
\midrule
CRAFT & 41 & 39 & \textbf{43} & 37 \\
\bottomrule
\end{tabular}
\vspace{0.8mm}
\begin{minipage}{0.97\linewidth}
\footnotesize
\textbf{Note.} Bold indicates the best EM.
$\theta{=}3$ is adopted in all main experiments.
\end{minipage}
\end{table}


\subsection{CRAFT Prompt Templates.}
\label{app:prompts}

The generator prompt (Box~\ref{box:gen}) is used across all CRAFT
iterations. On the first pass, the placeholder fields
\texttt{\$prev\_result\_block} and \texttt{\$feedback\_block} are empty;
on subsequent passes, they carry the prior output and verifier feedback,
respectively. The verifier prompt (Box~\ref{box:ver}) scores each
candidate on a 0--5 rubric; scores below 3 trigger re-generation with
actionable feedback.

\begin{tcolorbox}[title={\textbf{Box~\ref*{box:gen}: Generator Prompt}},
  label=box:gen, fontupper=\small, breakable, colback=white,
  colframe=black, colbacktitle=black!10, coltitle=black]

\texttt{Ignore previous conversations.}

\medskip
\textbf{TASK:}
Temporally order the provided list of adverse events based on their
sequence of appearance in the clinical notes and extract the most
specific mention for each adverse event from the notes.

\medskip
\textbf{Clinical Notes:} \textit{\{symptom\_text\}}

\textbf{Adverse Events to Temporally Order:} \textit{\{symptom\_list\}}

\medskip
\textbf{PROCESSING INSTRUCTIONS:}
\begin{itemize}[nosep,leftmargin=*]
  \item Start with the provided adverse events list and reorder it based
        on the sequence implied in the clinical notes.
  \item If several adverse effects are mentioned together without a clear
        temporal order, group them into the same block (inside a single
        dictionary).
  \item For each adverse event, extract the closest matching phrase from
        the clinical notes; if an adverse event is not mentioned, assign
        \texttt{"none"} as its value.
\end{itemize}

\medskip
\textbf{IMPORTANT RULES:}
\begin{itemize}[nosep,leftmargin=*]
  \item \textit{No Invention:} Never add, remove, or modify symptom
        names from the provided list.
  \item \textit{Mention Each Symptom Once:} Mention each symptom only
        once---at the first time it appears or becomes relevant.
  \item \textit{Specific vs.\ Generic Terms:} When a phrase matches both
        specific and generic symptoms, assign it to the specific term
        only. If ``Lip swelling'' or ``Pharyngeal swelling'' is matched,
        do NOT include the generic ``Swelling'' unless there is a
        separate, explicit mention of general swelling elsewhere.
  \item \textit{Temporal Evidence Required:} If multiple symptoms are
        mentioned across multiple sentences without clear timeline
        separation, group them together. Do NOT treat different sentences
        as different times unless there is an explicit temporal indicator
        (e.g., ``then,'' ``after that,'' ``later,'' specific dates or
        times).
  \item \textit{Unmentioned Symptoms:} Group all symptoms with no
        original mention together in a separate block with
        \texttt{"none"} as their value. This block should be placed
        after all the temporally ordered groups.
\end{itemize}

\medskip
\textbf{OUTPUT FORMAT:}
Return your answer only in valid JSON format---using a list of
dictionaries to represent temporal progression and grouping:

\begin{verbatim}
{[
  {"Erythema": ["redness in neck"]},
  {"Pain in extremity": ["sore arm"],
   "Pruritus": ["itchy feeling"]},
  {"Swelling": ["mild arm swelling"]}
]}
\end{verbatim}

\medskip
When revising, consider both the feedback (if any) and the previous
attempt result (if provided). Keep correct parts from prior attempts,
but fix issues based on feedback.

\medskip
\textit{\{prev\_result\_block\}} \quad \textit{\{feedback\_block\}}
\end{tcolorbox}

\begin{tcolorbox}[title={\textbf{Box~\ref*{box:ver}: Verifier Prompt}},
  label=box:ver, fontupper=\small, breakable, colback=white,
  colframe=black, colbacktitle=black!10, coltitle=black]

Given the original text and extracted symptoms below:

\medskip
\textbf{Original Text:} \textit{\{symptom\_text\}}

\textbf{Symptoms to Extract:} \textit{\{symptom\_list\}}

\textbf{Current Result (JSON):} \textit{\{initial\_result\}}

\medskip
\textbf{Scoring rubric} (0--5), +1 each if:
\begin{enumerate}[nosep,leftmargin=*]
  \item The JSON is valid and groups are earliest$\to$latest.
  \item Non-mentioned symptoms are presented as \texttt{"none"} in the
        last group.
  \item Every symptom in the list appears exactly once overall.
  \item Symptoms grouped together occur around the same time.
  \item Group ordering follows the text's temporal cues.
\end{enumerate}

\medskip
Return ONLY one of the following JSON objects:
\begin{itemize}[nosep,leftmargin=*]
  \item If score $\geq 3$:
        \verb|{"score": <int>, "feedback": ""}|
  \item If score $< 3$:
        \verb|{"score": <int>, "feedback": "<specific fixes>"}|
\end{itemize}
\end{tcolorbox}

\subsection{Heatmaps.}\label{appendix:heatmaps}

Figure~\ref{fig:before_after_comparison} compares ground-truth
before--after symptom relationships with the best-performing Claude CRAFT-Full
configuration across all three vaccine types and overall.

\subsection{Case Study 2: Verifier Miscalibration Induces Oscillation
(Claude Sonnet~4.5, GUIDE).}\label{appendix:case2}

\noindent\textbf{Symptom list:} Chest pain, Fatigue, Carditis,
Troponin increased

\smallskip
\noindent\textbf{Narrative (abridged):} \textit{``My son experienced
chest pain and was very tired --- Sat 6/19 [\ldots] Monday 6/21 at 11am
severe chest pain. Taken to hospital --- was hospitalized with heart
inflammation and very high troponin numbers until Thursday 6/24.''}

\smallskip
\noindent\textbf{Iteration 1} \hfill \textsc{Score: 2/5 --- below
$\theta$, continue}

\noindent\textbf{Verifier feedback:}
\begin{quote}
\texttt{ASPECT=GROUPING; OP=MERGE(g0, g1)}
\end{quote}

\noindent\textbf{Iteration 2} \hfill \textsc{Score: 2/5 --- below
$\theta$, continue}

\noindent\textbf{Verifier feedback:}
\begin{quote}
\texttt{ASPECT=GROUPING; OP=SPLIT(g0, move\_keys=[Chest pain, Fatigue],
to=NEW\_AFTER\_-1); OP=SPLIT(g1, move\_keys=[Carditis, Troponin
increased], \allowbreak\ to=NEW\_AFTER\_1)}
\end{quote}

\noindent\textbf{Iterations 3--4} \hfill \textsc{Score: 2/5 --- budget
exhausted, return last output}

\noindent\textbf{Verifier feedback:} \textit{(repeats $i{=}1$ and
$i{=}2$ feedback alternately)}

\smallskip
\noindent At $i{=}1$, Claude produces the exact gold-standard output,
correctly separating the pre-hospital symptoms (Chest pain, Fatigue)
from the hospitalization findings (Carditis, Troponin increased).
However, GUIDE's verifier assigns a score of 2 and instructs a
\textsc{Merge} operation --- because the narrative uses relative date
markers (``Sat 6/19'', ``Monday 6/21'') rather than explicit absolute
anchors, the verifier cannot confidently confirm distinct temporal
support for the two groups. The model faithfully follows the
instruction at $i{=}2$, merging all symptoms into a single stage and
producing a wrong output. The verifier then issues a \textsc{Split}
instruction, the model recovers the correct grouping at $i{=}3$, and
the cycle repeats. The loop terminates at $i{=}4$ on a merge step,
returning an incorrect final output despite the model having produced
the correct answer twice. This example illustrates how GUIDE's
verifier's strict requirement for explicit temporal anchors is
miscalibrated for narratives that express temporal order through
relative date references, causing it to reject a correct first-pass
output and drive the model into an unresolvable oscillation.
Table~\ref{tab:casestudy2} summarizes the predicted stages across
all four iterations.

\begin{table}[H]
\centering
\small
\setlength{\tabcolsep}{4pt}
\renewcommand{\arraystretch}{1.35}
\caption{Model outputs across iterations for Example~2
(Claude Sonnet~4.5, GUIDE). $\checkmark$ = matches gold;
$\times$ = grouping error.}
\label{tab:casestudy2}
\begin{tabular}{c c p{5.2cm}}
\toprule
\textbf{Iter} & \textbf{Score} & \textbf{Predicted Stages} \\
\midrule
Gold & --- &
  \{Chest pain, Fatigue\} $\to$
  \{Carditis, Troponin increased\} \\
\midrule
1 & 2/5 &
  \{Chest pain, Fatigue\} $\to$
  \{Carditis, Troponin increased\}
  \textcolor{green!60!black}{$\checkmark$} \\
\midrule
2 & 2/5 &
  \{Chest pain, Fatigue, \textbf{Carditis, Troponin increased}\}
  \textcolor{red}{$\times$} \\
\midrule
3 & 2/5 &
  \{Chest pain, Fatigue\} $\to$
  \{Carditis, Troponin increased\}
  \textcolor{green!60!black}{$\checkmark$} \\
\midrule
4 & 2/5 &
  \{Chest pain, Fatigue, \textbf{Carditis, Troponin increased}\}
  \textcolor{red}{$\times$} \\
\bottomrule
\end{tabular}
\end{table}

\end{document}